%% file: main.tex
\documentclass[10pt,twocolumn,letterpaper]{article}

\usepackage[pagenumbers]{cvpr} %

\usepackage{graphicx}
\usepackage{amsmath}
\usepackage{amssymb}
\usepackage{booktabs}

\usepackage[pagebackref,breaklinks,colorlinks]{hyperref}

\usepackage[capitalize]{cleveref}
\crefname{section}{Sec.}{Secs.}
\Crefname{section}{Section}{Sections}
\Crefname{table}{Table}{Tables}
\crefname{table}{Tab.}{Tabs.}

\usepackage{framed,multirow}
\RequirePackage{fix-cm}
\usepackage{enumitem}

\usepackage{amssymb}
\usepackage{latexsym}
\usepackage{amsmath}

\usepackage{url}
\usepackage{xcolor}
\definecolor{newcolor}{rgb}{.8,.349,.1}

\usepackage{hyperref}

\newcommand{\Fig}[1]{Fig.~\ref{fig:#1}}
\newcommand{\Figure}[1]{Figure~\ref{fig:#1}}

\newcommand{\Sec}[1]{Sec.~\ref{sec:#1}}

\begin{document}

\title{INVE: Interactive Neural Video Editing}%
\author{%
    Jiahui Huang$^{1,2}$\qquad
    Leonid Sigal$^{2,3,4,5}$\qquad
    Kwang Moo Yi$^{2}$\qquad
    Oliver Wang$^{1}$\qquad
    Joon Young Lee$^{1}$\\[.2in]
    $^1$Adobe Research\qquad 
    $^2$University of British Columbia\qquad 
    $^3$Vector Institute for AI \qquad \\
    $^4$CIFAR AI Chair \qquad
    $^5$NSERC CRC Chair \qquad
    \\[8pt]
    {\tt\small \href{https://gabriel-huang.github.io/inve/}{gabriel-huang.github.io/inve}}}

\maketitle
\input{0_abstract}

\input{front_figure.tex}
\input{1_introduction.tex}

\input{2_related.tex}

\input{3_method.tex}

\input{4_results.tex}
\input{5_conclusionn.tex}
\newpage
{
    \small
    \bibliographystyle{ieee_fullname}
    \bibliography{macros,main}
}
\end{document}

%% file: 0_abstract.tex
\begin{abstract}
We present Interactive Neural Video Editing (INVE), a real-time video editing solution, which can assist the video editing process by consistently propagating sparse frame edits to the entire video clip. 
Our method is inspired by the recent work on Layered Neural Atlas (LNA). LNA, however, suffers from two major drawbacks: (1) the method is too slow for interactive editing, and (2) it offers insufficient support for some editing use cases, including direct frame editing and rigid texture tracking.
To address these challenges we leverage and adopt highly efficient network architectures, powered by hash-grids encoding, to substantially improve processing speed. 
In addition, we learn bi-directional functions between image-atlas and introduce vectorized editing, which collectively enables a much greater variety of edits in both the atlas and the frames directly. 
Compared to LNA, our INVE reduces the learning and inference time by a factor of 5, and supports various video editing operations that LNA cannot. 
We showcase the superiority of INVE over LNA in interactive video editing through a comprehensive quantitative and qualitative analysis, highlighting its numerous advantages and improved performance.
A demo of our interactive editing interface can be found in the supplementary materials.

\end{abstract}

%% file: front_figure.tex
\begin{figure*}
    \centering
    \includegraphics[width=1\linewidth]{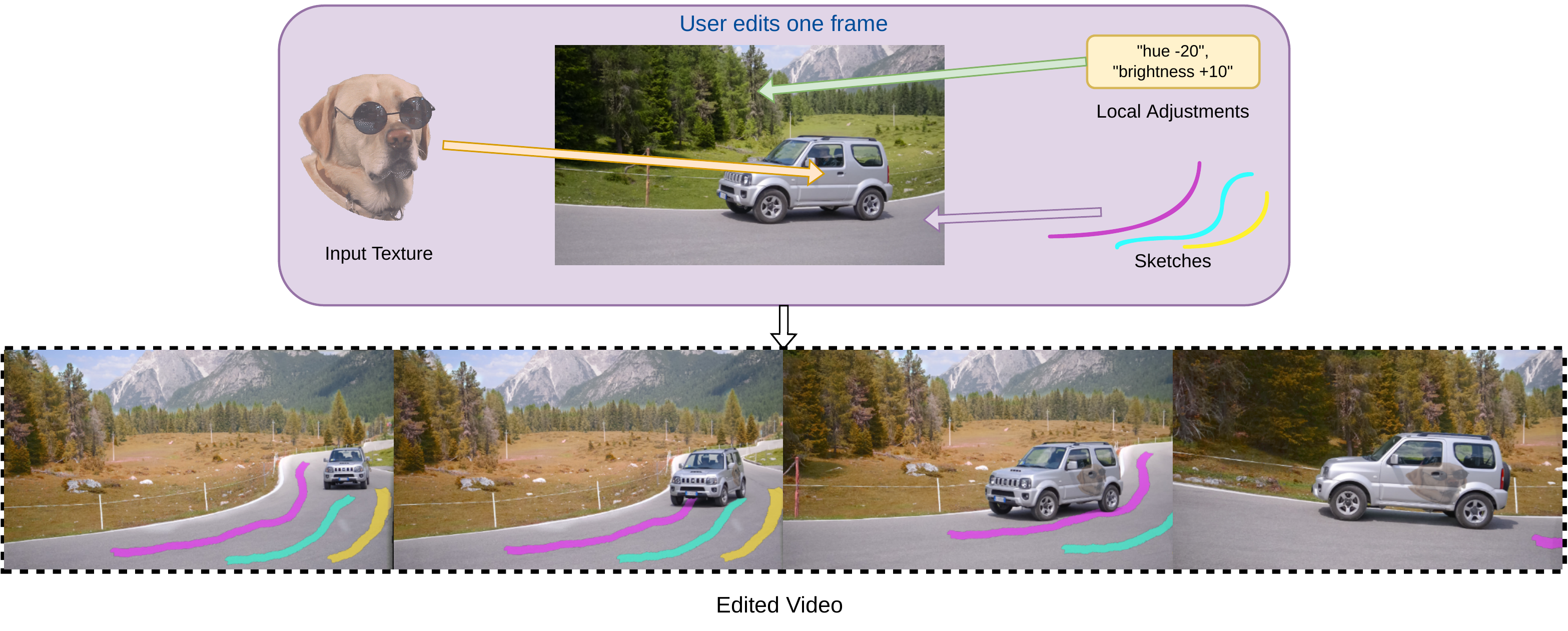}
    \caption{NeViE can propagate multiple types of image editing effects to the entire video in a consistent manner. In this case, the edits consist of (1) adding external graphics (dog picture) to the jeep; (2) Applying local adjustments (Hue -20, Brightness +10)) to the forest in the background; (3) Sketching on the road using the brush tool. All these types of edits can be propagated instantly from one frame to all other frames using the proposed approach.} 
    \label{fig:front_figure}
\end{figure*}

%% file: 1_introduction.tex
\section{introduction}
\label{sec:intro}

Image editing methods have progressed significantly over the years and now allow for complex editing operations with tools and interfaces that require little, to no, training for the novice user.
A testament to this is the availability and breadth of such tools in both commercial packages and mobile apps. 
Interactive {\em video} editing, on the other hand, remains a technical challenge that is yet to see similar success.  
At the same time, interactive video editing is an important application that can open a breadth of creative opportunities for non-professional users.
Such techniques can enable a variety of video editing operations, including local object/actor recoloring, relighting, texture editing, and many others. 

Progress in developing interactive video editing techniques has been slow due to the fundamental technical challenges that must be addressed before such techniques can become practical.
First, a scene being edited often consists of a non-static background and one-or-more foreground objects that undergo different motions. 
Edits must be localized and applied to these objects individually and then composed back to avoid unrealistic bleeding artifacts ({\em e.g.}, a ``dog" logo added to the foreground object (car) suddenly
sliding off and appearing in the background; see Fig~\ref{fig:front_figure}).
This requires robust temporally-consistent layered representations that must be learned in an unsupervised manner, which in itself is a challenging task for realistic scenes.
Second, asking
the user to edit each frame individually is both unrealistic and impractical from the user effort point of view.
Further, inconsistencies that may result from independent frame-based editing tend to have glaring visual artifacts as humans are very sensitive to temporal inconsistencies.
As a result, a mechanism for sparse editing in time (and possibly in space) and an automated way to propagate such edits are useful features of a video editor.
Third, the creative process of video editing often assumes some level of interactive control over the edits.
Building an approach that takes minutes or hours to apply an edit would significantly stifle the creativity of the user and render such techniques practically undesirable.

Earlier 2D approaches advocated keyframe editing directly in the frames and propagated these edits using frame-to-frame tracking ({\em e.g.}, using optical flow) \cite{Chen2017CoherentOV,Huang2016TemporallyCC}.
Such approaches tend to be challenged by drift and occlusions, producing artifacts that highly depend on the video content, selected keyframes, and the edits applied. 
Recently developed {\em layered neural atlas} representations~\cite{Kasten2021LayeredNA}, enables {\em consistent} editing of videos, containing arbitrary types of moving objects or background, by representing the video by a set of layered neural 2D atlases ({\em i.e.}, images), one for each object and one for background.
Such representations have a number of appealing properties, which include locality and consistency of edits enabled by editing in the individual atlases as opposed to the keyframes directly. 
However, certain challenges remain. 
First, the estimated mapping from the atlas to video pixels is not bijective, enabling edits {\em only} in the atlas.
This is less ideal for certain applications, as typically non-linear mapping (represented by a neural network), makes it difficult to anticipate how a specific atlas edit will be perceived in the video.
This results in less than intuitive editing and potential unexpected deformation artifacts. 
Second, current {\em layered neural atlas} representations tend to be slow to compute, making the editing effectively non-interactive. 

In this paper, our focus is on addressing these core challenges, while, at the same time, building on the successes of neural atlas representations.
We do this by proposing to learn a bi-directional mapping between the atlases and the image, along with vectorized sketching that enables us to make consistent edits either in the atlas itself or in the image (by back-projecting the edits onto the learned atlas).
This significantly extends the editing operations available to the user.
Further, we adopt and develop multi-resolution hash coding \cite{Mller2022InstantNG} to the task of layered neural atlas representations, which significantly improves both the learning and inference speed allowing more interactive user interactions and control. 

\vspace{0.1in}
\noindent 
{\bf Contributions:} Our contributions are both technical / methodological as well as user-centric -- enabling richer vocabulary of consistent and interactive video edits for novice users. We summarized our contributions below:
\vspace{-.5em}
\begin{itemize}[leftmargin=*]
    \item INVE achieves $5\times$ faster training and inference speed compared to existing methods~\cite{Kasten2021LayeredNA};
    \item we introduce inverse mapping to enable rigid texture tracking effects;
    \item we support editing multiple video effects \textit{independently} via layered editing;
    \item we introduce Vectorized Sketching for artifact-free sketch editing at the frame level.
\end{itemize}

%% file: 2_related.tex
\input{pipeline.tex}

\section{Related Works}
\subsection{Video Effects Editing}
Video effects editing 
involves adding or modifying visual effects in a video. 
Many methods have been proposed in the literature to address this problem, including both traditional and deep learning-based approaches.
One traditional approach is to use keyframes to represent the effects and interpolate between them to generate a video with smooth transitions \cite{Huang2016TemporallyCC}.
Deep learning-based methods have also been explored for video effects editing. For example, Generative Adversarial Networks (GANs) \cite{Goodfellow2014GenerativeAN} have been used to generate new video frames with different visual effects, such as removing rain or snow \cite{Tang2018RemovalOV}, generating a photorealistic video from an input segmentation map video \cite{Wang2018VideotoVideoS}, or generating frames with controlled, plausible motion \cite{Huang2021LayeredCV}.
In addition, other deep learning-based methods have been used for video effects editing, such as video style transfer \cite{Jamriska2019StylizingVB}, which involves transferring the style of one or few keyframes to the entire video, super-resolution \cite{Shi2016RealTimeSI}, which involves increasing the resolution of a video.
In our work, we focus on propagating single-frame edits to the entire video in a consistent manner, where videos can be edited as if editing a single image, we demonstrate that our editing pipeline can propagate multiple types of image editing effects to the entire video consistently. 

\subsection{Video Propagation}
Video propagation is an important area of research in computer vision, which focuses on the propagation of visual information through time in video data. Some methods \cite{Chen2017CoherentOV,Huang2016TemporallyCC} purpose to propagate information based on constraints posed by optical flow, however, since optical flow is only computed within neighboring frames, these methods often suffer from propagation drifting over a long period of time. 
Deep learning-based methods \cite{Jabri2020SpaceTimeCA,Jampani2016VideoPN,Oh2018FastVO,Wang2019LearningCF,Xu2022TemporallyCS}, have also been extensively explored in recent years. For example, Video Propagation Networks \cite{Jampani2016VideoPN} first splats information to a bilateral space, then uses a learned filter to slice the information back to image space. 
Some other approaches \cite{Kasten2021LayeredNA,RavAcha2008UnwrapMA} learn unwarped 2D texture maps, then edits can be performed on these maps, and be warped back to all frames. For example, Layered Neural Atlases (LNA) decomposes the input video into the foreground and background layers, and learns two mapping networks that map each video pixel to the UV coordinates on the fore-background texture maps, which they call atlases. Our method is conceptually similar to LNA, except that we made several improvements to the edit-ability and overall editing experience (including learning and inference speed). 

\subsection{Implicit Neural Representation}
Recent works have shown that implicit neural representation can be very robust for representing visual data. For example, representing 3D geometry with neural radiance fields \cite{Barron2021MipNeRFAM,Mildenhall2020NeRFRS,Tancik2022BlockNeRFSL,Xu2022PointNeRFPN}, representing 2D image data for image compression \cite{Dupont2021COINCW}, image super-resolution \cite{Chen2020LearningCI}, and image generation \cite{Anokhin2020ImageGW,Skorokhodov2020AdversarialGO}. Representing 3D video volume using implicit functions has also been explored, for example, Mai {\em et al.} proposed Motion-Adjustable Neural Implicit Video Representation \cite{Mai2022MotionAdjustableNI}, which allows re-synthesizing videos with different motion properties, and Layered Neural Atlases \cite{Kasten2021LayeredNA}, which enables consistent video editing.
Meanwhile, highly efficient network architectures \cite{Mller2021TinnyCUDA} have been purposed to reduce the computational cost of training and testing of these implicit networks, and hashed encoding \cite{Mller2022InstantNG} was purposed to drastically improve the convergence speed of training such networks.
In our work, we represent an input video with six implicit neural networks: two forward mapping networks, two backward mapping networks, one opacity network, and one atlas network, all implemented with high-efficiency network architectures and encoding functions. With these networks combined, our approach enables interactive and consistent editing, as well as basic point tracking on videos.

%% file: pipeline.tex
\begin{figure*}
    \centering
    \includegraphics[width=1\linewidth]{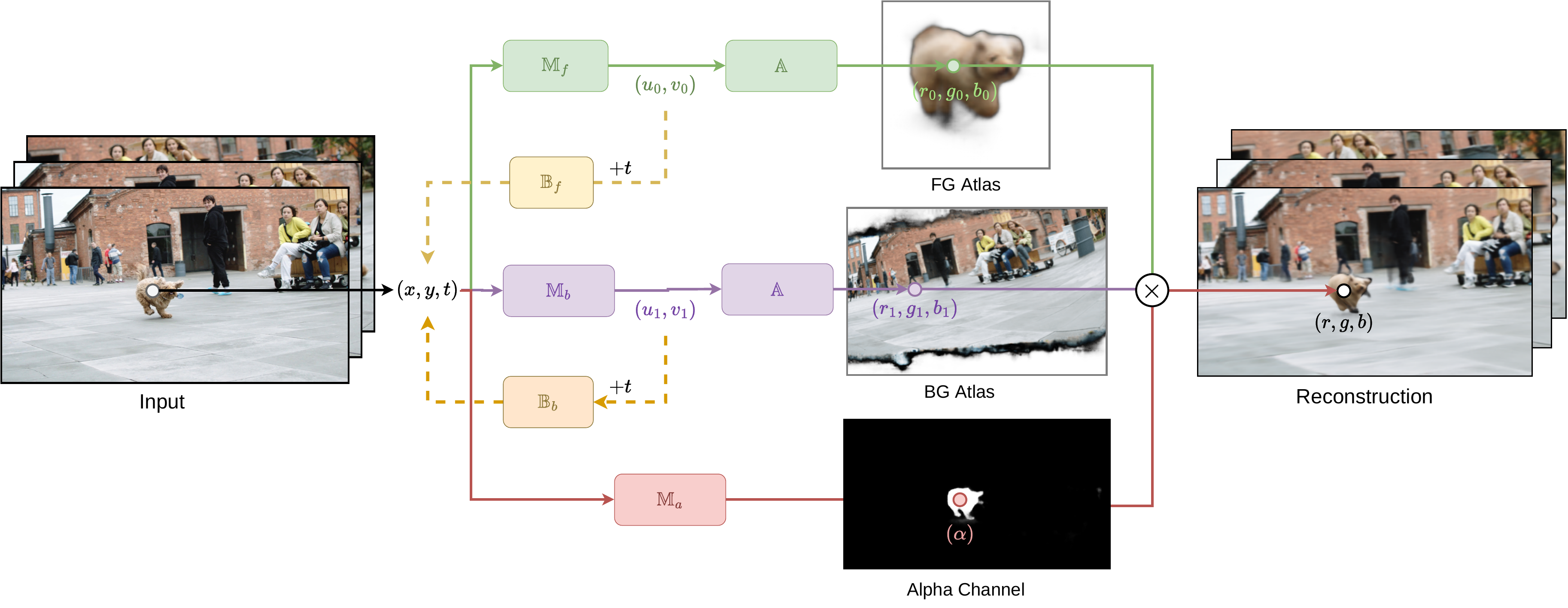}
    \caption{
    Our forward mapping pipeline (\textit{solid lines}) closely follows LNA's approach. Each video pixel location $(x, y, t)$ is fed into two mapping networks, $\mathbb{M}_f, \mathbb{M}_b$ to predict $(u, v)$ coordinates on each atlas. Then these coordinates are fed into the atlas network $\mathbb{A}$ to predict the RGB color on that atlas. Finally, we use the opacity value $\alpha$ predicted by the alpha network $\mathbb{M}_a$ to compose the reconstructed color at location $(x,y,t)$. Our backward mapping pipeline (\textit{dotted lines}) maps atlas coordinates to video coordinates, it takes an $(u, v)$ coordinate, as well as the target frame index $t$ as input, and predicts the pixel location $(x,y,t)$. With the forward and backward pipelines combined, we can achieve long-range point tracking on videos.
    }
    \label{fig:pipeline} 
\end{figure*}

%% file: 3_method.tex
\section{Interactive Neural Video Editing (INVE)}

In this section, we describe our method for interactive neural video editing, INVE.
As noted in \Sec{intro}, our focus is to perform edits directly on a given frame, which is then automatically propagated to all other frames consistently.
To explain our method, we first review Layered Neural Atlases~\cite{Kasten2021LayeredNA} in \Sec{review}, which is the base framework that we build our method on top of.
We then discuss how we achieve interactive performance by boosting computation speed in \Sec{boosted_speed}, then discuss how we enable rigid texture tracking -- a critical feature for easy video editing, by introducing inverse mapping in \Sec{inverse_mapping}.
Lastly, we discuss how we edit videos with our method, with a focus on vectorized sketching that allows artifact-free sketch editing at the frame level in \Sec{vectorized_sketching}.

\subsection{Review of Layered Neural Atlases}
\label{sec:review}
Layered Neural Atlases (LNA)~\cite{Kasten2021LayeredNA} represents a video sequence with three sets of neural networks: (1) the mapping networks, which we write as $\mathbb{M}: (x, y, t) \rightarrow (u, v)$ that map 3D video pixel coordinates to 2D texture coordinates on the atlases; (2) the atlas networks, $\mathbb{A}(u, v) \rightarrow (r, g, b)$, which predict the color of a given texture coordinate on a given atlas; (3) the opacity network, $\mathbb{O}(x, y, t) \rightarrow \alpha$, that predicts the opacity values at each pixel w.r.t. each atlas.
Each of the above networks is represented by a coordinate-based MLP.

The entire framework is trained end-to-end in a self-supervised manner. The main loss is an unsupervised reconstruction loss, where the network is tasked to reconstruct the RGB color of a given video pixel location.
LNA also has three regularization losses: (1) \textit{Rigidity loss}: encourages the mapping from video pixels to the atlas to be locally rigid; (2) \textit{Consistency loss}: encourages corresponding pixels in consecutive frames to be mapped at the same location on the atlases, it uses pre-computed optical flow to estimate the pixel correspondence.
(3) \textit{Sparsity loss}: encourages the atlases to contain minimal content needed to reconstruct the video. 

Once the neural representation (the atlas) for the video is obtained via training, video editing is performed by editing directly on the atlases.
These `atlas-level edits' are then mapped to each frame by the learned mapping function.
The final edited video is obtained by blending these edits with the original video. 
Hence, this atlas is in fact an intermediate layer that eventually needs to be mapped onto each frame to be actually realized.
Thus, while it is possible to visually inspect the atlas, edits on this atlas are not how an edit would look when mapped onto an actual frame, making it suboptimal for performing video editing.
Moreover, mapping in LNA is \textit{unidirectional} -- from the frames to the atlas, which makes it difficult for any intuition for point/texture tracking.

In LNA, the authors briefly mentioned that a user can edit the video by directly sketching on the frame, this is realized by first generating a frame edit layer (a raster image of the size of the frame) containing all the sketches, then mapping this frame edit layer to the atlas edit layer (this is done by mapping every pixel on the frame edit layer to the atlas), and finally interpolate the color of the atlas edit layer.
Doing so has two obvious drawbacks, first, mapping the entire frame edit layer is computationally expensive, the total mapping cost is $H \times W$ pixels, and second, the resulting atlas edit layer may contain undesirable artifices stemming from the interpolation (see \Figure{vector_compare}).
\input{converge_speed}

\subsection{Boosted Training \& Inference Speed}
\label{sec:boosted_speed}
Besides the problem of mapping and editing of atlas, another important issue with LNA is that it is too slow for interactive video editing. 
We make an observation that the task of atlas-based video modeling is similar, at the core, to the task of gigapixel image approximation. Specifically, they both use implicit neural representations to ``memorize'' the input data. 
LNA uses sinusoidal positional encoding~\cite{Tancik2020FourierFL} to increase the frequency of the network input, which shifted all the ``memorization" overload to the subsequent MLPs. 
To tackle this problem we turn our attention to recent neural field backbones that utilize multiresolution hash grids (InstantNGP)~\cite{Mller2022InstantNG}.
In our pipeline, instead of the sinusoidal positional encoding, we opt for the multiresolution hash grid, which shared part of the ``memorization'' overload from the MLPs to the trainable encoding itself, this can lead to a significant boost in convergence speed.
Furthermore, we use a GPU parallelized and fully fused MLP implementation using the TinyCUDA library~\cite{Mller2021TinnyCUDA}
that significantly improves the computation speed of our pipeline.
We further train significantly fewer iterations than LNA, which we detail in \Sec{impl}.

\subsection{Inverse Mapping for point tracking on videos}
\label{sec:inverse_mapping}
As noted earlier, LNA only supports one directional mapping, from frame coordinates to atlas coordinates--we refer to this as forward mapping.:
\begin{equation}
    \mathbb{M}(x, y, t) \rightarrow (u,v)
    \;.
\label{eq:forwardmap}
\end{equation}
Editing using LNA's pipeline is achieved by sampling the edited color from the atlas layers, this is equivalent to warping from the atlas plane to the frame plane using a dense warping field, defined by an \textit{untrackable} inverse mapping function, which can result in undesirable warping deformations for rigid texture tracking.

Conversely, in our work, we propose to also model the inverse mapping function using neural networks.
Specifically, we introduce additional mapping networks (one per layer) on top of the LNA framework that map from atlases to frames. 
Formally, given a point $(u, v)$ on the atlas, and the destination frame index $t$, the inverse mapping function $\mathbb{B}$ will predict the landing pixel coordinate $(x, y)$ on frame $t$:
\begin{equation}
    \mathbb{B}(u, v, t) \rightarrow (x, y, t)
    \;.
\label{eq:reversemap}
\end{equation}
In this way, given a point $\mathsf{p}$ on frame $t$, we can easily track its trajectory $\mathbf{P}$ by first mapping it to the atlas using forward mapping $\mathbb{M}$, then use the inverse mapping to calculate its corresponding locations on the rest of the frames, that is:
\begin{equation}
    \mathbf{P} = \mathbb{B}(u, v, T)
    \;.
\end{equation}
Where $T = \{t_0, t_1, .., t_N\}$, indicating the frame index.

The training of the inverse mapping networks is supervised by the forward mapping networks. 
After fully training the forward mapping networks, we start training the inverse mapping by randomly sampling the video to obtain pixel--atlas coordinate pairs using forward mapping.
We then use these paired data to train the inverse mapping networks. 
As we desire to be able to predict \emph{all} frames that the $(u,v)$ coordinate maps to, we extend the input domain with the frame time, as seen in in Equation~\ref{eq:reversemap}.

\subsection{Layered Editing}
\label{sec:layered_editing}
Image editing is usually done with layers.
For example, in Adobe Photoshop, users can overlay multiple editable layers on top of the original image, and each layer can be accessed and edited individually.
The final output is usually a back-to-front composition of all layers.
We adopt a similar idea for our editing pipeline, we overlay three editable layers on top of the atlases, and each one of them stores a different type of edit, so that they can be accessed individually should one wish to do so. 
Specifically:
\begin{itemize}[leftmargin=*]
    
\item {\bf Sketch edits.}
A user can draw vectorized sketches using the brush tool (see more on \Sec{vectorized_sketching}).

\item {\bf Texture edits.} 
When the user ``draws'' an imported asset (this is done by clicking on the frame/atlas to set the anchor point and dragging to set the size), the anchor point coordinates and the size of the texture (width and height) will be stored, and the texture will be ``pasted'' onto the texture edit layer in the atlas space.

\item {\bf Metadata edits.}
A user can perform local adjustments ({\em i.e.}, increase the brightness) at any desired region on the frame by drawing out these regions with the brush tool, the adjustment metadata will be carried by the brush stroke, and stored in the metadata edit layer in the atlas space.
\end{itemize}
A user can edit \textit{directly} on those layers, or edit on the frames. When editing on frames, edits are first mapped to atlas coordinates, then stored in the corresponding layer depending on the edit type.

The final result is rendered pixel-by-pixel.
For each video pixel, we first map its coordinate to its atlas coordinate using the forward mapping function, we then look up the edits of that pixel in the atlas space, and finally, we render the RGB value of that pixel by using back-to-front composition through all edits and the original pixel value.
\input{vectorize.tex}
\subsection{Vectorized Sketching}
\label{sec:vectorized_sketching}
Being able to sketch directly on frames is a very desirable function in video editing, for example, performing free-form annotations when analysing a sports video. 
As mention earlier in \Sec{review}, frame sketch editing using LNA's pipeline is sub-optimal due to its slowness and undesirable artifacts.
These artifacts arise due to the fact that the atlas has to be \emph{resampled} onto the target image domain for rendering. If the sampling rate of the atlas is too low, we can see aliasing artifacts in the rendering (see \Fig{vector_compare}).

To address these two problems, we propose vectorized sketching (\Fig{vectorize}), where we represent a user sketch as a continuous vectorized representation, so that we can avoid resampling it.
We choose to represent the sketch as a polygonal chain, which is defined by a sequence of $K$ control points:
\begin{equation}
    \mathcal{E}_f = \left\{(x_{i-1}, y_{i-1}):(x_i, y_i)\right\}, i\in\{1,2,...K\}
    \;.
\end{equation}
We then map these control points to atlas coordinates,
\begin{equation}
    (u_i, v_i) = \mathbb{M}(x_i, y_i), i\in\{1,2,...K\}
    \;,
\end{equation}
then define the polygonal chain in the atlas space as:
\begin{equation}
    \mathcal{E}_a = \left\{(u_{i-1}, v_{i-1}):(u_i, v_i)\right\}, i\in\{1,2,...K\}
    \;.
\end{equation}
By doing so, we can avoid warping artifacts and bring down the mapping cost from $H \times W$ pixels to $K$ pixels.

In addition, vectorized sketches can carry additional attributes other than color alone.
For example, in our editing pipeline, each sketch stroke can carry a metadata field, which includes brightness, hue and saturation values.
These can be used to apply local adjustments as discussed earlier in \Sec{layered_editing}.

\subsection{Implementation Details}
\label{sec:impl}

\paragraph{Early Stopping.}

In our work, the main aim is to perform video editing, not creating a neural representation for videos.
Hence, as long as we have accurate mappings between the atlas and the frames, the quality of the atlas and the reconstructed video frames are irrelevant.
Thus, we train our method only until the mapping network matures, which we empirically found to be much quicker than the atlas network $\mathbb{A}$ of our pipeline.

\paragraph{Details.}
Our implementation of the Neural Video editing pipeline closely follows Layered Neural Atlases (LNA) \cite{Kasten2021LayeredNA}.
As in LNA, we train and test our method on videos consisting of 70 frames with resolution of 768 $\times$ 432.
We randomly sample 10,000 video pixels per batch and train the model for around 12,000 iterations, which is notably less than the LNA implementation (300,000 iterations). 
In total, our model has $\sim$1.7 M parameters, and 
requires 5 GB GPU memory. 
Training our model takes about 5 minutes, and rendering the final video takes 2.8s ($\sim$25 fps) on an NVIDIA RTX 4090 GPU.

%% file: converge_speed.tex
\begin{figure}
    \centering
    \includegraphics[width=1\linewidth,trim = 0 0 0 10, clip]{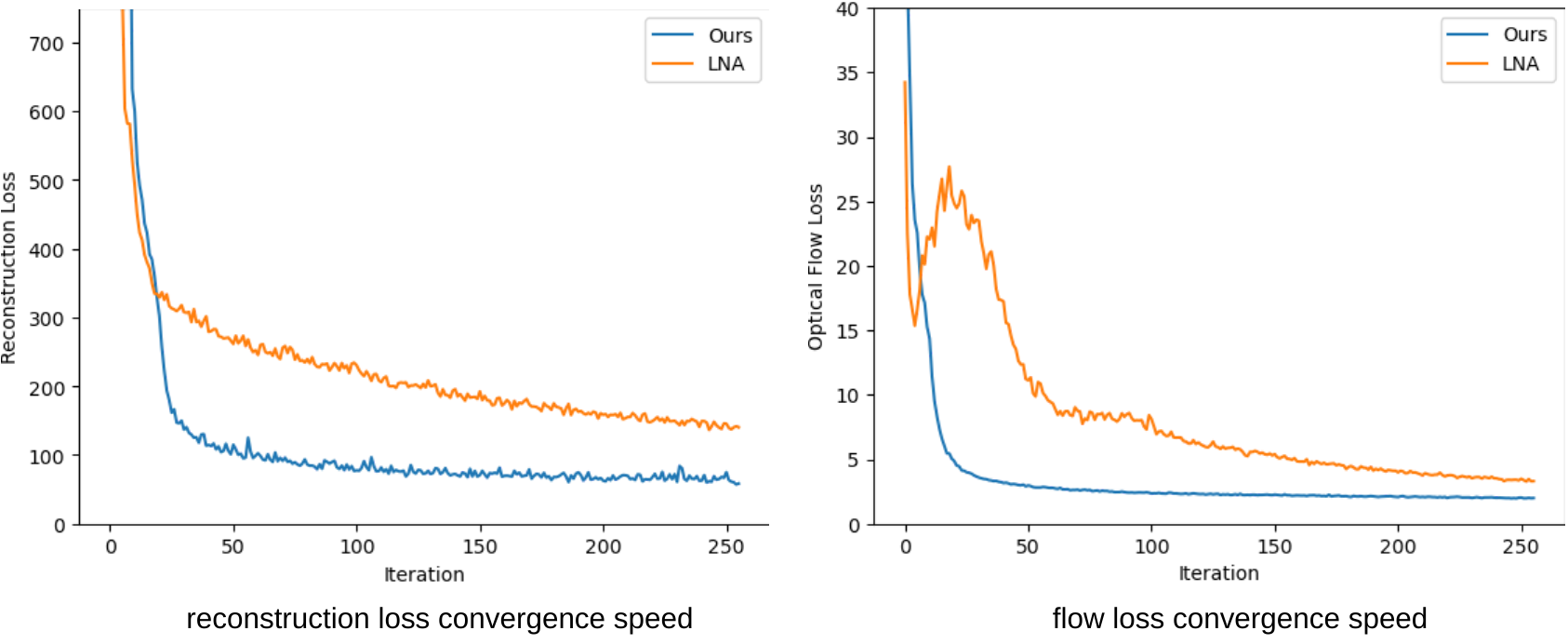}
    \caption{Convergence Speed Comparison. Given the same number of training iterations, both reconstruction quality (measured by the reconstruction loss) and mapping accuracy (measured by the optical flow loss) of our model converges faster than LNA's. }
    \label{fig:converge_speed}
\end{figure}

%% file: vectorize.tex
\begin{figure}
    \centering
    \includegraphics[width=0.98\linewidth]{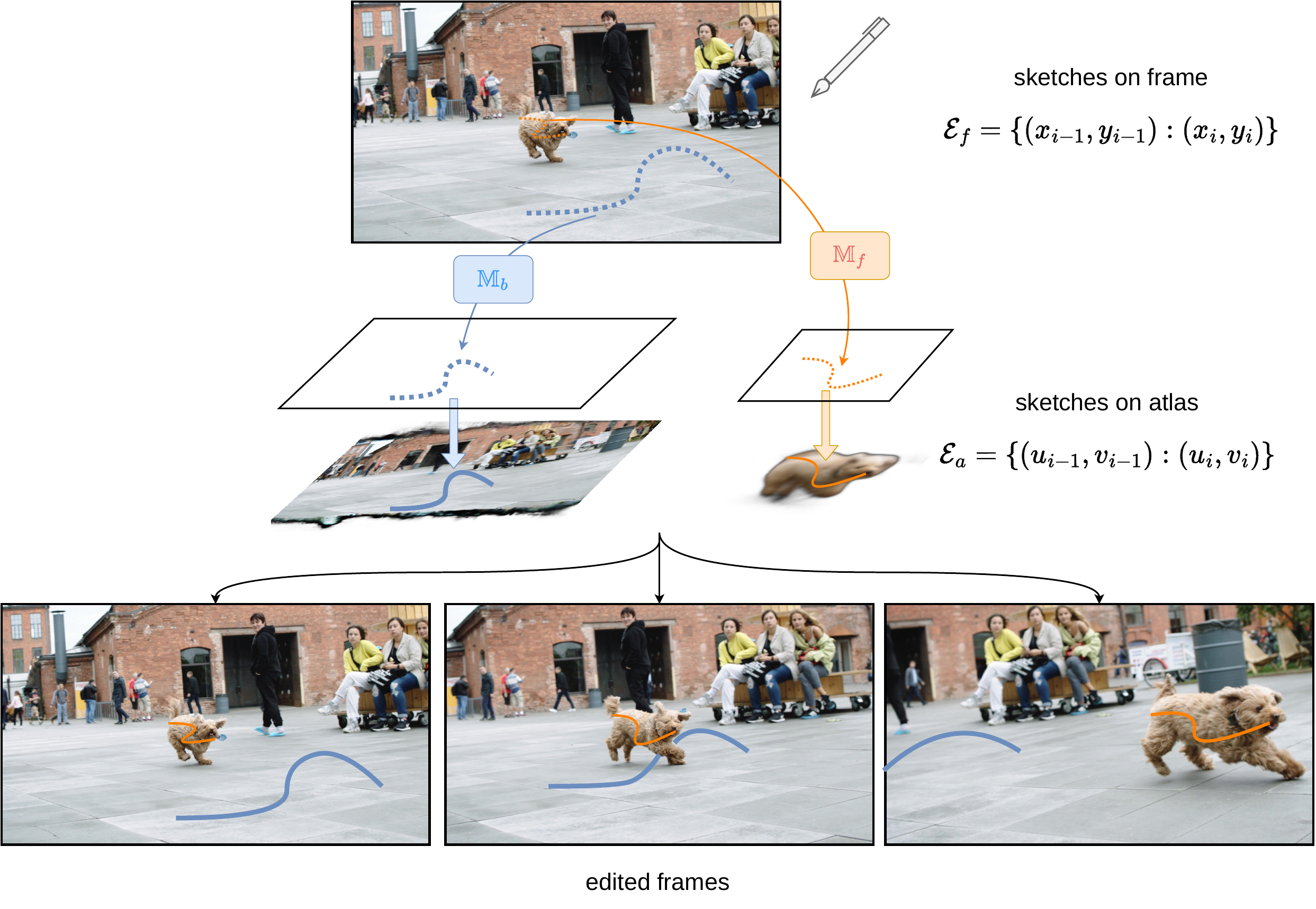}
    \caption{
    Vectoriezed Sketching. User sketches directly on the frame, the mouse tracks $\left\{(x_{i}, y_{i})\right\}$ that define these sketches will be mapped to atlas coordinates $\left\{(u_{i}, v_{i})\right\}$, then these tracks will be used to render polylines on the atlas edit layer.
    }
    \label{fig:vectorize} 
\end{figure}

%% file: 4_results.tex
\input{vector_compare}
\input{rec_speed_compare}

\section{Results}
In this section, we evaluate the effectiveness of our proposed method on videos from the DAVIS dataset \cite{Pont-Tuset_arXiv_2017}, as well as our own videos. Following the approach of LNA, we utilize RAFT \cite{Teed2020RAFTRA} for optical flow extraction. We discovered that the quality of the masks significantly impacts the reconstruction results and convergence speed. Therefore, we opted for a more precise mask extractor \cite{Oh2019VideoOS} instead of MaskRCNN \cite{He2017MaskR}. Our approach aims to improve two critical aspects of LNA: training / testing speed, and edit-ability. We conduct all our experiments on a single NVIDIA RTX 4090 GPU.

\subsection{Improved Training \& Inference Speed}
To improve training and testing speed, we first adapt the GPU-optimized Fully Fused MLP \cite{Mller2021TinnyCUDA} architecture into our pipeline, which significantly increased the computation speed per sample batch, from 23 iterations (10,000 samples/batch) per second to 48 iterations. We further improved the convergence speed of our model by adapting the multiresolution hash encoding \cite{Mller2022InstantNG}, as shown in \Figure{converge_speed}, after training the same number of iterations, both the reconstruction loss (representing reconstruction quality) and the flow loss (representing mapping accuracy) converges faster on our model than LNA. On \Figure{rec_speed_compare}, we show that given the same training time, the quality of reconstructed frames from our model is much better than LNA's both visually and quantitatively (see PSNR on the bottom of each image). At test time, the rendering speed of our approach is 24.81 FPS, compared to LNA's 5.34 FPS. The boost in both training and inference speed makes our method more favorable for interactive video editing.
\input{backward_compare}

\subsection{Inverse Mapping for Point Tracking}
The LNA approach only supports one directional forward mapping. Editing using forward mapping alone is equivalent to warping the edited texture using a dense warping field, which can be insufficient to support many editing applications, such as adding rigid textures that track a single/few points. For example, \Figure{backward_compare} shows a case where the user wants to add a ``hat'' texture to the dancer. If the video is edited using LNA's one-directional mapping, the hat texture needs to be warped to the frame using the dense warping field defined by the forward mapping function (see top row), as a result, the texture is warped completely out of shape (see the bottom row). With our inverse mapping function, the user can add the texture that tracks a point on her head, which gives more promising results (see middle row).
\input{layered_editing}
\input{results}

\subsection{Layered Editing Pipeline}
Our layered editing pipeline allows  users to overlay multiple editable layers on top of the atlases, and each layer can be accessed and edited individually. 
On \Figure{layered_editing}, we demonstrate the results of all three types of edits supported by our pipeline. On the top row, we show that user sketches can be consistently propagated to all frames in the video. In the middle row, we show that the user can apply local adjustments (in this case, lower saturation and higher brightness) to a specific region in the scene by using our vectorized sketching tool, which can carry the adjustment metadata field, and on the bottom row, we show that user can import external graphic textures that track and deform with the moving foreground object.
On \Figure{results}, we showcase some videos edited using our pipeline; our method can propagate various types of edits consistently to all frames.

\subsection{Vectorized Sketching} 
Our purposed vectorized sketching allows us to map the polygonal chains (represented by a set of control points) that define the sketch strokes directly to the atlases, which can help reduce computational cost, and avoid artifacts stemming from LNA's frame editing pipeline (map frame sketches as a raster image). On \Figure{vector_compare}, we show the resulting edited atlas produced by vectorized sketching (left), LNA editing using linear interpolation (middle), and LNA editing using nearest neighbor interpolation (right). One can easily observe that mapping frame sketches using our method provides a continuous sketch stroke with consistent color, whereas LNA's pipleine either produces non-continuous sketch, or inconsistent color, depending on the interpolation method. 

%% file: vector_compare.tex
\begin{figure}
    \centering
    \includegraphics[width=1\linewidth]{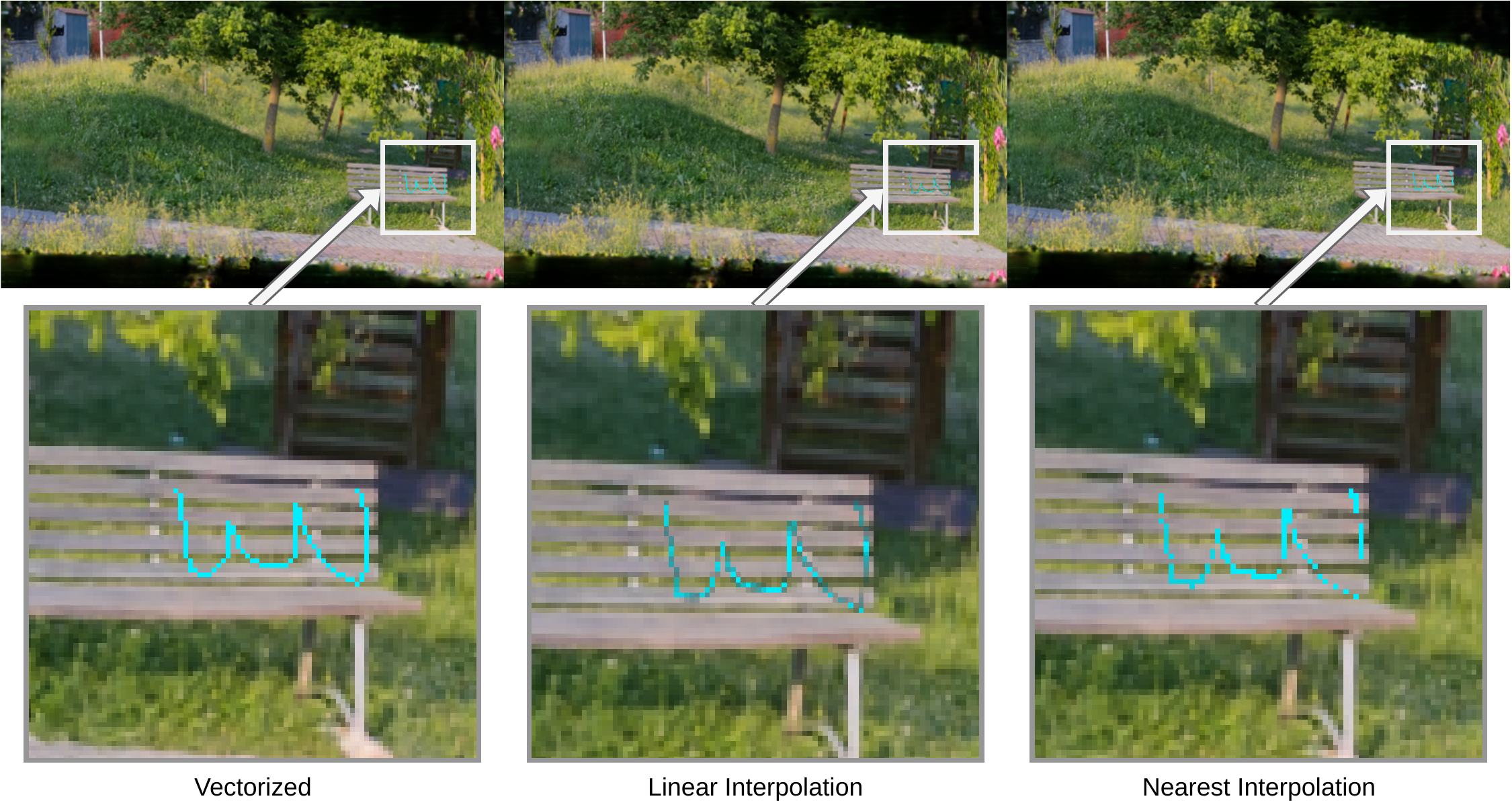}
    \caption{Our vectorized sketching allows users to perform sketch editing \textit{directly} on frames free from resampling artifacts (left), whereas frame editing using LNA's pipeline either results in inconsistent color (middle) or noncontinuous sketches (right).}
    \label{fig:vector_compare}
\end{figure}

%% file: rec_speed_compare.tex
\begin{figure*}
    \centering
    \includegraphics[width=\linewidth]{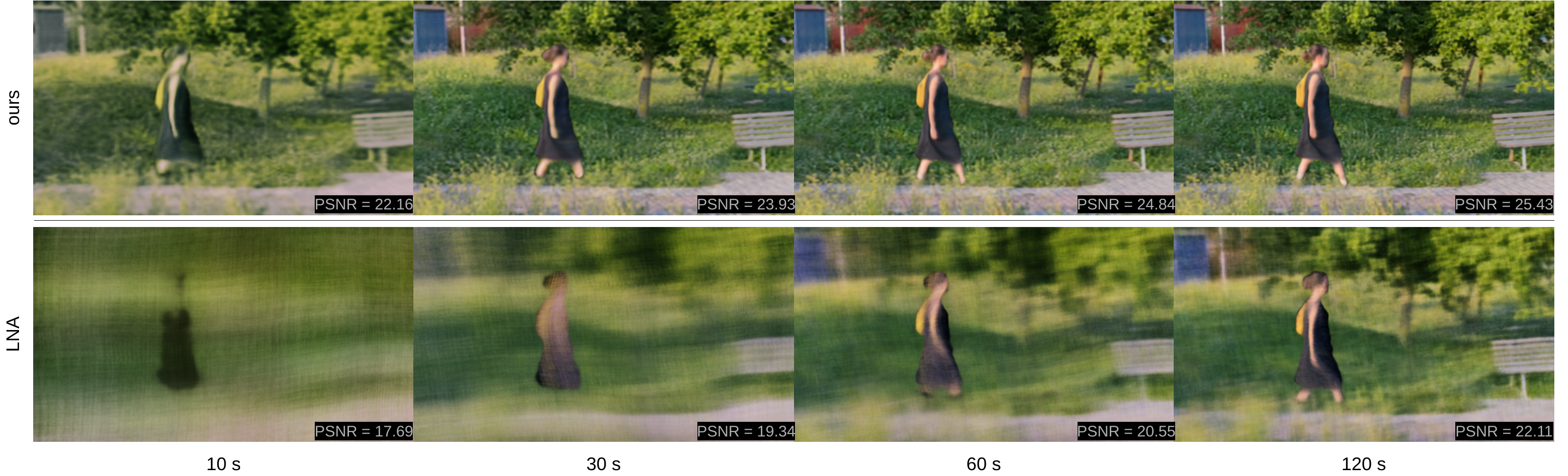}
    \caption{Given the same training time, the quality of reconstructed frames produced by our model is much better than LNA’s both visually and quantitatively (see PSNR onthe bottom of each image).}
    \label{fig:rec_speed_compare} 

\end{figure*}

%% file: backward_compare.tex
\begin{figure*}
    \centering
    \includegraphics[width=1\linewidth]{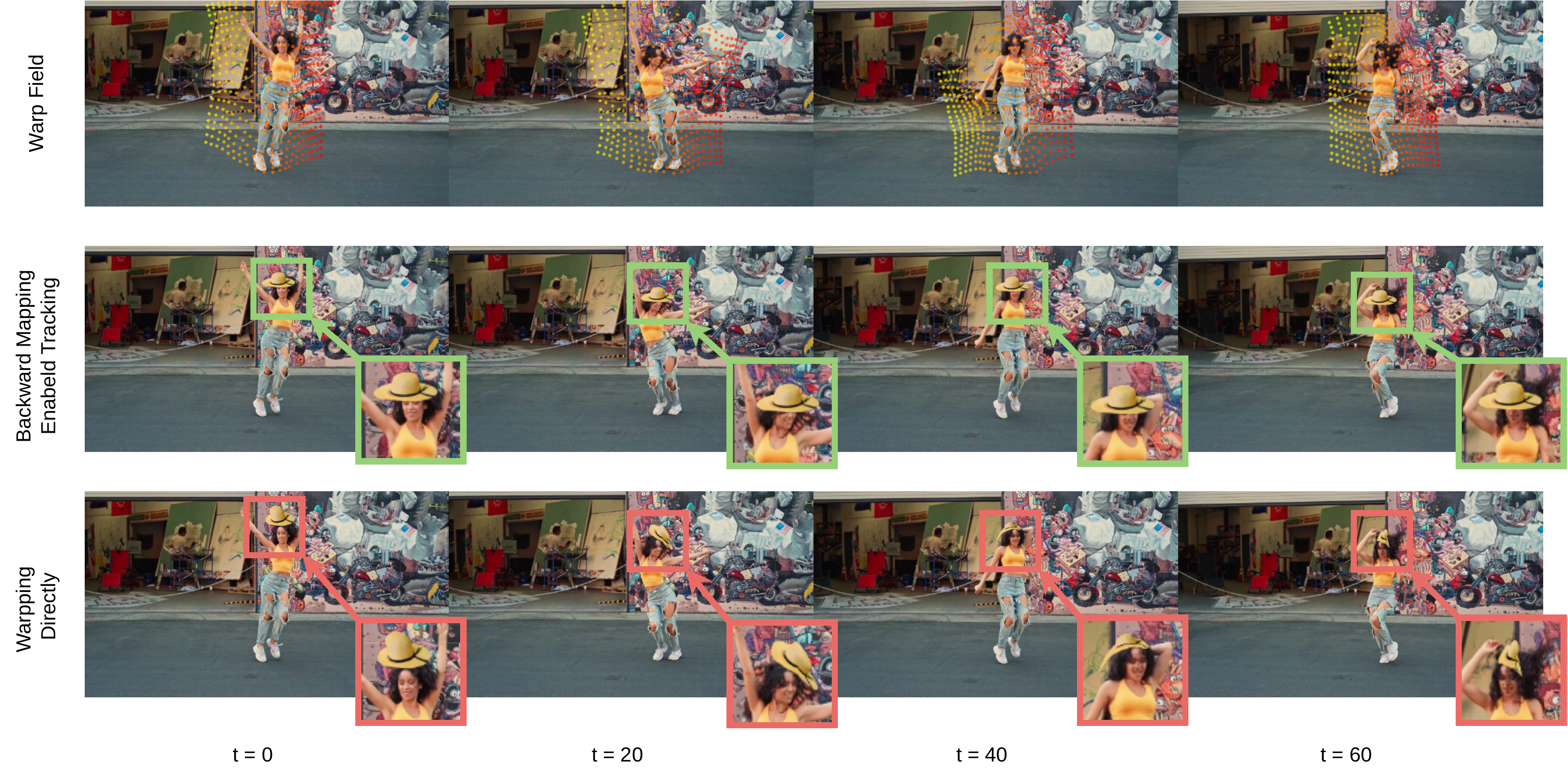}
    \caption{Inverse Mapping enabled tracking. Editing using LNA's forward mapping alone is equivalent to warping the edited texture using a dense warping field (visualized on top row), which can lead to undesired warpping effects (bottom row). Our approach introduces inverse mapping, which enables video particle tracking spamming all frames, here we showcase using tracking function to insert a texture that tracks a selected point (middle row). }
    \label{fig:backward_compare}
\end{figure*}

%% file: layered_editing.tex
\begin{figure*}
    \centering
    \includegraphics[width=0.95\linewidth]{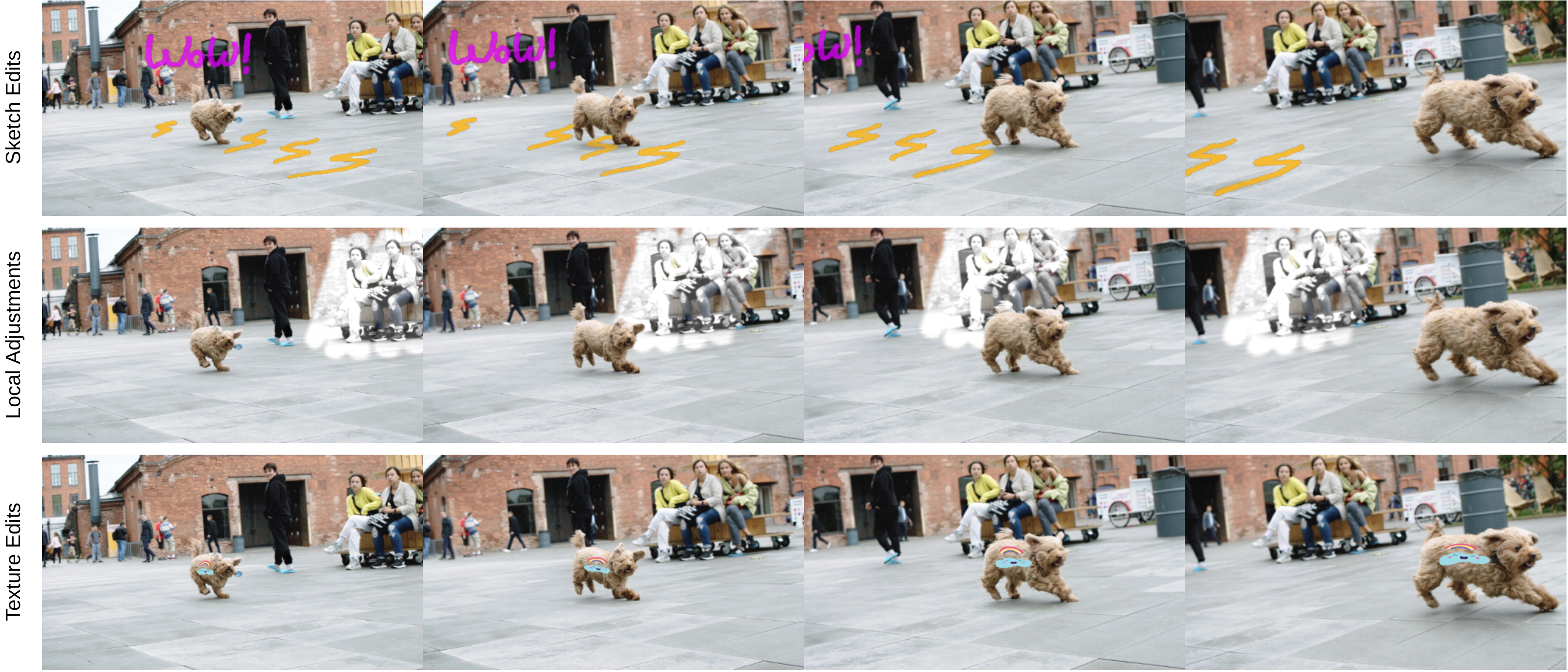}
    \caption{Layered Editing. Our layered editing pipeline supports three types of edits: 1) Sketch Edits (top), where users can sketch scribbles using the brush tool; 2) Local Adjustments (middle), users can apply local adjustments (brightness, saturation, hue) to a specific region in the scene; 3) Texture Edits (bottom), users can import external graphics that tracks and deforms with the moving object.}
    \label{fig:layered_editing} 
\end{figure*}

%% file: results.tex
\begin{figure*}
    \centering
    \includegraphics[width=\linewidth]{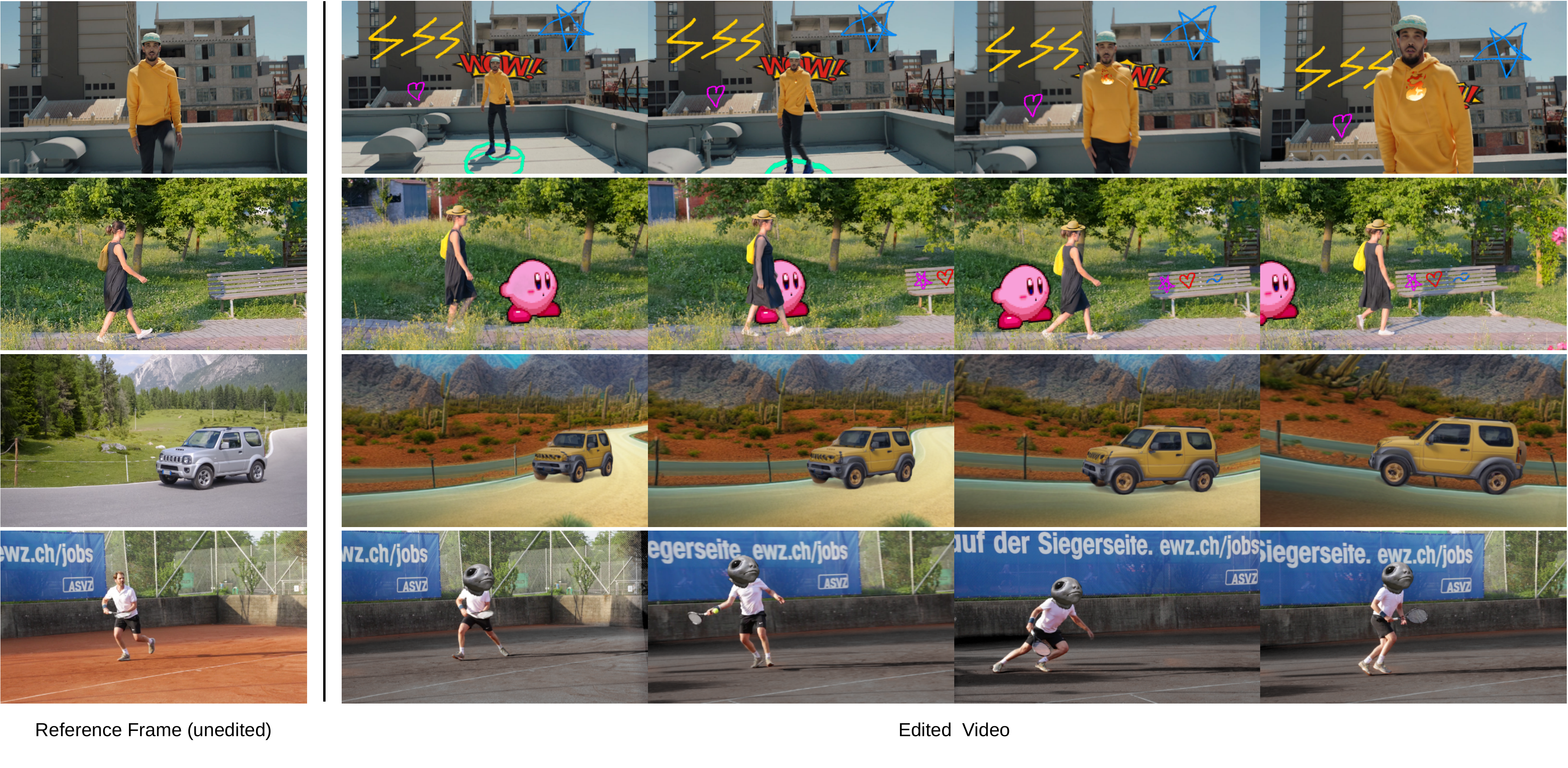}
    \caption{
    Results showcase. Here we showcase some videos edited using our pipeline, on the left is a reference of an unedited frame, and on the right are the sampled frames from the edited video.
    }
    \label{fig:results} 
\end{figure*}

%% file: 5_conclusionn.tex
\section{Conclusion.}
We propose INVE: Interactive Neural Video Editing, an interactive video editing pipeline, which makes video editing easier and more accessible by instantly and consistently propagating single-frame edits to the entire video. Our method is inspired by the recent work Layered Neural Atlas (LNA), upon which we made several improvements in speed and in editability. We believe that INVE can significantly improve the video editing experience, particularly for beginners who may be intimidated by the complexity of traditional editing tools.

%% file: main.bbl
\begin{thebibliography}{10}\itemsep=-1pt

\bibitem{Anokhin2020ImageGW}
Ivan Anokhin, Kirill~V. Demochkin, Taras Khakhulin, Gleb Sterkin, Victor~S.
  Lempitsky, and Denis Korzhenkov.
\newblock Image generators with conditionally-independent pixel synthesis.
\newblock {\em 2021 IEEE/CVF Conference on Computer Vision and Pattern
  Recognition (CVPR)}, pages 14273--14282, 2020.

\bibitem{Barron2021MipNeRFAM}
Jonathan~T. Barron, Ben Mildenhall, Matthew Tancik, Peter Hedman, Ricardo
  Martin-Brualla, and Pratul~P. Srinivasan.
\newblock Mip-nerf: A multiscale representation for anti-aliasing neural
  radiance fields.
\newblock {\em 2021 IEEE/CVF International Conference on Computer Vision
  (ICCV)}, pages 5835--5844, 2021.

\bibitem{Chen2017CoherentOV}
Dongdong Chen, Jing Liao, Lu Yuan, Nenghai Yu, and Gang Hua.
\newblock Coherent online video style transfer.
\newblock {\em 2017 IEEE International Conference on Computer Vision (ICCV)},
  pages 1114--1123, 2017.

\bibitem{Chen2020LearningCI}
Yinbo Chen, Sifei Liu, and Xiaolong Wang.
\newblock Learning continuous image representation with local implicit image
  function.
\newblock {\em 2021 IEEE/CVF Conference on Computer Vision and Pattern
  Recognition (CVPR)}, pages 8624--8634, 2020.

\bibitem{Dupont2021COINCW}
Emilien Dupont, Adam Goli'nski, Milad Alizadeh, Yee~Whye Teh, and A. Doucet.
\newblock Coin: Compression with implicit neural representations.
\newblock {\em ArXiv}, abs/2103.03123, 2021.

\bibitem{Goodfellow2014GenerativeAN}
Ian~J. Goodfellow, Jean Pouget-Abadie, Mehdi Mirza, Bing Xu, David
  Warde-Farley, Sherjil Ozair, Aaron~C. Courville, and Yoshua Bengio.
\newblock Generative adversarial nets.
\newblock In {\em NIPS}, 2014.

\bibitem{He2017MaskR}
Kaiming He, Georgia Gkioxari, Piotr Doll{\'a}r, and Ross~B. Girshick.
\newblock Mask r-cnn.
\newblock {\em IEEE Transactions on Pattern Analysis and Machine Intelligence},
  42:386--397, 2017.

\bibitem{Huang2021LayeredCV}
Jiahui Huang, Yuhe Jin, Kwang~Moo Yi, and Leonid Sigal.
\newblock Layered controllable video generation.
\newblock {\em European Conference on Computer Vision (ECCV)}, 2022.

\bibitem{Huang2016TemporallyCC}
Jia-Bin Huang, Sing~Bing Kang, Narendra Ahuja, and Johannes Kopf.
\newblock Temporally coherent completion of dynamic video.
\newblock {\em ACM Transactions on Graphics (TOG)}, 35:1--11, 2016.

\bibitem{Jabri2020SpaceTimeCA}
A. Jabri, Andrew Owens, and Alexei~A. Efros.
\newblock Space-time correspondence as a contrastive random walk.
\newblock {\em ArXiv}, abs/2006.14613, 2020.

\bibitem{Jampani2016VideoPN}
V. Jampani, Raghudeep Gadde, and Peter Gehler.
\newblock Video propagation networks.
\newblock {\em 2017 IEEE Conference on Computer Vision and Pattern Recognition
  (CVPR)}, pages 3154--3164, 2016.

\bibitem{Jamriska2019StylizingVB}
Ondrej Jamriska, S{\'a}rka Sochorov{\'a}, Ondrej Texler, Michal Luk{\'a}c,
  Jakub Fiser, Jingwan Lu, Eli Shechtman, and Daniel S{\'y}kora.
\newblock Stylizing video by example.
\newblock {\em ACM Transactions on Graphics (TOG)}, 38:1--11, 2019.

\bibitem{Kasten2021LayeredNA}
Yoni Kasten, Dolev Ofri, Oliver Wang, and Tali Dekel.
\newblock Layered neural atlases for consistent video editing.
\newblock {\em ACM Transactions on Graphics (TOG)}, 40:1--12, 2021.

\bibitem{Mai2022MotionAdjustableNI}
Long Mai and Feng Liu.
\newblock Motion-adjustable neural implicit video representation.
\newblock {\em 2022 IEEE/CVF Conference on Computer Vision and Pattern
  Recognition (CVPR)}, pages 10728--10737, 2022.

\bibitem{Mildenhall2020NeRFRS}
Ben Mildenhall, Pratul~P. Srinivasan, Matthew Tancik, Jonathan~T. Barron, Ravi
  Ramamoorthi, and Ren Ng.
\newblock Nerf: Representing scenes as neural radiance fields for view
  synthesis.
\newblock In {\em European Conference on Computer Vision}, 2020.

\bibitem{Mller2022InstantNG}
Thomas M{\"u}ller, Alex Evans, Christoph Schied, and Alexander Keller.
\newblock Instant neural graphics primitives with a multiresolution hash
  encoding.
\newblock {\em ACM Transactions on Graphics (TOG)}, 41:1--15, 2022.

\bibitem{Mller2021TinnyCUDA}
Thomas M{\"u}ller, Fabrice Rousselle, Jan Nov'ak, and Alexander Keller.
\newblock Real-time neural radiance caching for path tracing.
\newblock {\em ACM Transactions on Graphics (TOG)}, 40:1--16, 2021.

\bibitem{Oh2018FastVO}
Seoung~Wug Oh, Joon-Young Lee, Kalyan Sunkavalli, and Seon~Joo Kim.
\newblock Fast video object segmentation by reference-guided mask propagation.
\newblock {\em 2018 IEEE/CVF Conference on Computer Vision and Pattern
  Recognition}, pages 7376--7385, 2018.

\bibitem{Oh2019VideoOS}
Seoung~Wug Oh, Joon-Young Lee, N. Xu, and Seon~Joo Kim.
\newblock Video object segmentation using space-time memory networks.
\newblock {\em 2019 IEEE/CVF International Conference on Computer Vision
  (ICCV)}, pages 9225--9234, 2019.

\bibitem{Pont-Tuset_arXiv_2017}
Jordi Pont-Tuset, Federico Perazzi, Sergi Caelles, Pablo Arbel\'aez, Alexander
  Sorkine-Hornung, and Luc {Van Gool}.
\newblock The 2017 davis challenge on video object segmentation.
\newblock {\em arXiv:1704.00675}, 2017.

\bibitem{RavAcha2008UnwrapMA}
Alex Rav-Acha, Pushmeet Kohli, Carsten Rother, and Andrew~William Fitzgibbon.
\newblock Unwrap mosaics: a new representation for video editing.
\newblock {\em ACM SIGGRAPH 2008 papers}, 2008.

\bibitem{Shi2016RealTimeSI}
Wenzhe Shi, Jose Caballero, Ferenc Husz{\'a}r, Johannes Totz, Andrew~P. Aitken,
  Rob Bishop, Daniel Rueckert, and Zehan Wang.
\newblock Real-time single image and video super-resolution using an efficient
  sub-pixel convolutional neural network.
\newblock {\em 2016 IEEE Conference on Computer Vision and Pattern Recognition
  (CVPR)}, pages 1874--1883, 2016.

\bibitem{Skorokhodov2020AdversarialGO}
Ivan Skorokhodov, Savva Ignatyev, and Mohamed Elhoseiny.
\newblock Adversarial generation of continuous images.
\newblock {\em 2021 IEEE/CVF Conference on Computer Vision and Pattern
  Recognition (CVPR)}, pages 10748--10759, 2020.

\bibitem{Tancik2022BlockNeRFSL}
Matthew Tancik, Vincent Casser, Xinchen Yan, Sabeek Pradhan, Ben Mildenhall,
  Pratul~P. Srinivasan, Jonathan~T. Barron, and Henrik Kretzschmar.
\newblock Block-nerf: Scalable large scene neural view synthesis.
\newblock {\em 2022 IEEE/CVF Conference on Computer Vision and Pattern
  Recognition (CVPR)}, pages 8238--8248, 2022.

\bibitem{Tancik2020FourierFL}
Matthew Tancik, Pratul~P. Srinivasan, Ben Mildenhall, Sara Fridovich-Keil,
  Nithin Raghavan, Utkarsh Singhal, Ravi Ramamoorthi, Jonathan~T. Barron, and
  Ren Ng.
\newblock Fourier features let networks learn high frequency functions in low
  dimensional domains.
\newblock {\em ArXiv}, abs/2006.10739, 2020.

\bibitem{Tang2018RemovalOV}
Lai~Meng Tang, Li~Hong Lim, and Paul Siebert.
\newblock Removal of visual disruption caused by rain using cycle-consistent
  generative adversarial networks.
\newblock In {\em ECCV Workshops}, 2018.

\bibitem{Teed2020RAFTRA}
Zachary Teed and Jia Deng.
\newblock Raft: Recurrent all-pairs field transforms for optical flow.
\newblock {\em ArXiv}, abs/2003.12039, 2020.

\bibitem{Wang2018VideotoVideoS}
Ting-Chun Wang, Ming-Yu Liu, Jun-Yan Zhu, Guilin Liu, Andrew Tao, Jan Kautz,
  and Bryan Catanzaro.
\newblock Video-to-video synthesis.
\newblock In {\em Neural Information Processing Systems}, 2018.

\bibitem{Wang2019LearningCF}
X. Wang, A. Jabri, and Alexei~A. Efros.
\newblock Learning correspondence from the cycle-consistency of time.
\newblock {\em 2019 IEEE/CVF Conference on Computer Vision and Pattern
  Recognition (CVPR)}, pages 2561--2571, 2019.

\bibitem{Xu2022PointNeRFPN}
Qiangeng Xu, Zexiang Xu, Julien Philip, Sai Bi, Zhixin Shu, Kalyan Sunkavalli,
  and Ulrich Neumann.
\newblock Point-nerf: Point-based neural radiance fields.
\newblock {\em 2022 IEEE/CVF Conference on Computer Vision and Pattern
  Recognition (CVPR)}, pages 5428--5438, 2022.

\bibitem{Xu2022TemporallyCS}
Yi Xu, Badour Albahar, and Jia-Bin Huang.
\newblock Temporally consistent semantic video editing.
\newblock In {\em European Conference on Computer Vision}, 2022.

\end{thebibliography}
